\documentclass{llncs}
\usepackage{makeidx}  

\usepackage{amssymb}
\usepackage{graphicx}

\usepackage{epstopdf}
\usepackage[dvips]{epsfig}
\usepackage{tikz}
\usetikzlibrary{positioning}
\usepackage{natbib}
\bibpunct{[}{]}{;}{a}{,}{,}

\usepackage{url}

\begin{document}

\mainmatter 

\title{Predicting health inspection results from online restaurant reviews}

\titlerunning{Predicting health inspection results from online restaurant reviews}

\author{Samantha Wong\inst{1} \and Hamidreza Chinaei\inst{1} \and Frank Rudzicz\inst{2}}

\institute{The G. Raymond Chang School of Continuing Education, Ryerson University,\\ Toronto, ON, Canada\\
\email{samantha.y.wong@ryerson.ca, hrchinaei@ryerson.ca}
\\ \and
Department of Computer Science, University of Toronto,\\ Toronto, ON, Canada\\
\email{frank@cs.toronto.edu }
}

\maketitle

\begin{abstract}
Informatics around public health are increasingly shifting from the professional to the public spheres. In this work, we apply linguistic analytics to restaurant reviews, from Yelp, in order to automatically predict official health inspection reports. We consider  two types of feature sets, i.e., keyword detection and topic model features, and use these in several classification methods. Our empirical analysis shows that these extracted features can predict public health inspection reports with over 90\% accuracy using simple support vector machines.
\end{abstract}

\section{Introduction}
Typically, official Health and Safety inspections of restaurants are only conducted one to three times a year, and involve fixed checklists related to food safety\footnote{\url{http://chd.region.waterloo.on.ca/en/healthylivinghealthprotection/resources/sample_restaurant_inspection.pdf}}. Meanwhile, consumer reviews of these restaurants are posted publicly on sites like Yelp and TripAdvisor much more frequently. Naturally, this leads to the question of whether restaurant reviews can support the public health inspection process by flagging potentially high risk restaurants that may need more frequent inspections.

 To address this question, we compare official restaurant health inspection results from the Kitchener-Waterloo region against text comments found in Yelp reviews of the relevant restaurants. In particular, we extract series of keywords and topic model features and apply these to a few classification methods. Our extracted features can predict the public health inspection with 90\% accuracy when used with a support vector machine (SVM).

\begin{figure}[!ht]
\centering
\includegraphics[scale=.3]{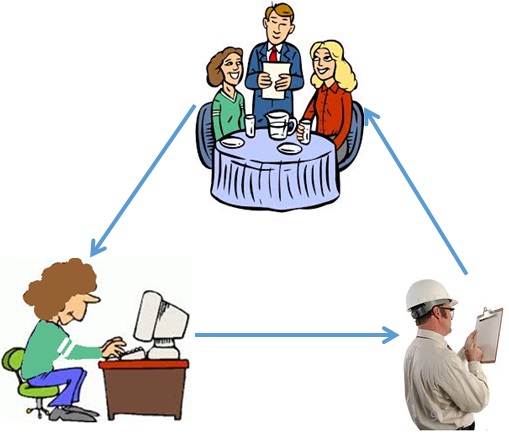}
\caption{Can text restaurant reviews predict health inspection results?}
\end{figure}

\subsection{Related Work}
\cite{kang2013where} matched keywords and phrases (unigram and bigram) from text restaurant reviews and other Yelp metadata for to results from the Department of Health's research results, with the goal of allocating inspectors more efficiently. That study used liblinear's SVM with L1 regularization and 10-fold cross-validation, which was able to achieve 82.68\% accuracy when classifying text reviews with problematic restaurants. Kang {\em et al.} suggested that the highest accuracy can be achieved using only contentful unigrams and bigrams. They also presented cue features for different topics/categories such as Hygienic (e.g., {\em gross}, {\em mess}, {\em sticky}, {\em smell}, {\em dirty}), Cuisines ({\em Vietnamese}, {\em Dim Sum}, {\em Thai}), and Sentiment ({\em cheap}, {\em never}). Our work extends theirs in a few ways, including our use of topic modelling.

\cite{Blei2003} proposed  a Bayesian topic modeling approach called latent dirichlet allocation (LDA)  which is used
for discovering hidden topics of documents. In this model, a document can be represented as a mixture of the learned hidden topics, where each hidden topic is represented by a distribution over words occurred in the document.  The authors also applied their LDA method to text classification on Reuters-21578 dataset. They trained a SVM on the topic model representations applied by LDA and compared this SVM to an SVM trained on all the word features. They concluded that the former SVM can potentially produce better accuracy results. In this work, we use both topic model features and keyword features in an SVM.

\cite{sahami1998bayesian} studied junk mail filtering using na\"{i}ve Bayes and message text, along with several other non-textual features. They also described building a corpus of keywords from the text of junk emails, and how some of those keywords were removed as part of feature selection based loosely on Zipf's law. They were able to achieve 97.1\% precision and 94.3\% recall for classifying junk mail. In our work, we use also na\"{i}ve Bayes and text reviews, but for a different purpose. 

\cite{nsoesie2014online} addressed the spread of food-borne illness through Yelp reviews using keyword selection (according to symptoms) and clustering to predict {\em the geographical location} where a food-borne outbreak originated. They also highlighted limitations in using Yelp data to support health-related studies, specifically with regards to review bias, and issues related to timing. Nonetheless, Nsoesie {\em et al.} confirmed that it is possible to crowd-source food and health reports, given careful analysis.

Other work has applied analysis of textual signals in social media to public health surveillance more generally, especially with regards to the spread of disease \citep{aramaki2011twitter,sadilek2012modeling,sadilek2013nemesis,lamb2013separating,dredze2013carmen,von2010assessment}. The increase of recent work in this area suggests at a shift towards `crowd-sourced' public health informatics.

\section{Data}
In our study, we use  Yelp\footnote{\url{https://www.yelp.ca/dataset_challenge/dataset}} and Waterloo Public Health (WPH)\footnote{\url{http://www.regionofwaterloo.ca/en/regionalGovernment/FoodPremiseDataset.asp}} data. We extract facility information from WPH, as well as official inspection results (date, type, and recommended actions). From Yelp, we extract business information (matched with WPH facility information) and tokenize text reviews. These tokens are used in a document/term matrix, and  act as  features in this work. The other set of features are learned topic models, i.e., the distribution of topics for each learned topic. We use binary prediction labels: ``action'' indicating non-compliance or other major problem with the restaurant, and ``no action'' if the restaurant passes health inspection.



      

\begin{figure}[!b]
\centering
\includegraphics[scale=.39]{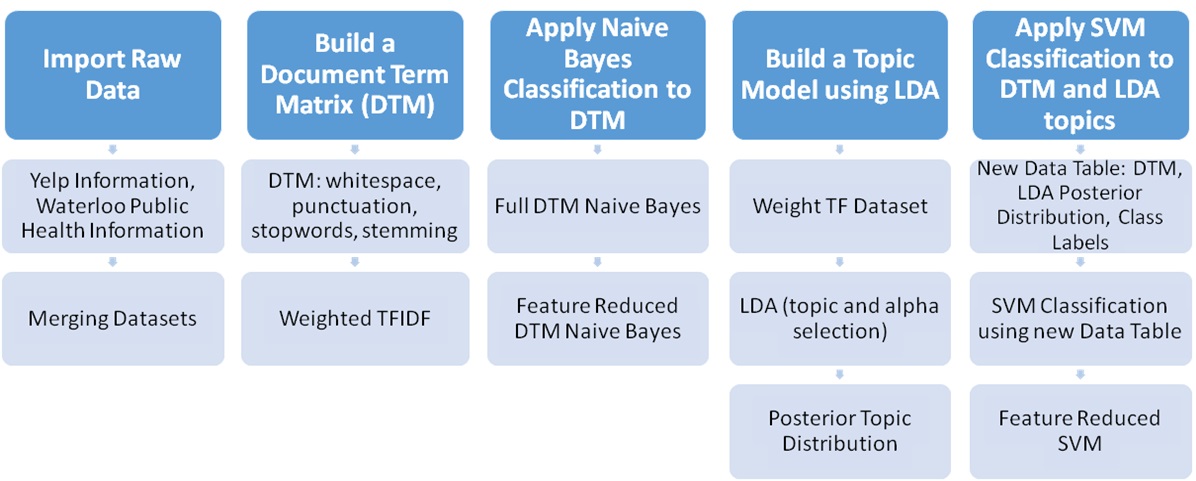}
\caption{High-level approach to automatic, crowd-sourced health inspection method.} \label{fig:approach}
\end{figure}

\section{Approach}

Figure~\ref{fig:approach} summarizes our approach, summarized here.

\paragraph{Import row data} First, we filter business data to those in the Kitchener-Waterloo-Cambridge area. We then extract  the data for the region from WPH. Finally, we merge the two datasets, resulting in a list of 126 restaurants, their Yelp Business IDs, and WPH Facility IDs. This allows us to merge WPH health inspection results and textual Yelp reviews. A preliminary qualitative analysis of documents labeled with ``action'' suggested that relatively {\em few} of the associated reviews included negative terms (e.g., {\em dirty}, {\em sick}, {\em bad}). Interestingly, many of those reviews focused on poor service, food temperature, or the restaurant environment generally, rather than direct references to public health issues.

\paragraph{Build a document-term matrix (DTM)}
We build a {\em tf-idf} weighted document-term matrix using all text reviews. First, we create a regular document-term matrix with plain text (stripped of whitespaces, punctuation, stopwords, and stems). Then, we weigh each term with {\em tf-idf}. Specifically, we calculate term frequencies for each word using a function based on: \[\mbox{ TF($t$,$d$) } = \frac{\mbox{ Number of times term $t$ appears in a document $d$} } {\mbox{Total number of terms in document $d$ } }. \] We then calculate the  inverse document frequencies using: \[\mbox{IDF($t$,$D$)} = \log \left( \frac{\mbox{ Total number of documents in set $D$}} {\mbox{Number of documents in $D$ containing term $t$}} \right).\] Finally, the overall score for term-document pair $(t,d)$ is $TF(t,d)\cdot IDF(t,D)$.

We use the built-in functionality of the \texttt{tm}  package in R\footnote{\url{http://www.inside-r.org/packages/cran/tm/docs/weightTfIdf}}, which includes further optimizations for {\em tf-idf}, and we use the resulting matrix for all of our calculations below.

\paragraph{Applying na\"{i}ve Bayes (NB) classification to the DTM}
We apply NB to the full DTM created above but obtained poor results. NB had difficulty classifying ``No Action'' reviews, with only 11.78\% accuracy and 10\% sensitivity.

Therefore, we reduce the number of features in the DTM by using only the $N$ most frequent few contentful terms, identified by sums of the weighted values for each column of the DTM. Optimizing this process gives $N=300$, with which we achieve an accuracy of 47.13\% and a sensitivity of 13.22\%, which is still sub-optimal.

In an effort to improve the results, we add latent Dirichlet analysis (LDA)~\citep{Blei2003} to perform topic modelling, and incorporate these topics into a SVM classifier.

\paragraph{Build a topic model using LDA}
First, we apply {\em term frequency weighting} to the original document term matrix (as opposed to {\em tf-idf}), prior to LDA. We empirically select 20 topics and, similar to previous work~\citep{Amit2007}, we set the Dirichlet prior on the per-document topic distributions to $\alpha=1+50/k$ (where $k$ is the number of topics), giving $\alpha = 3.5$. Topics appeared to be mostly divided by the type of cuisine (e.g., {\em Chinese}, {\em wings}, {\em burgers}, {\em tea/coffee}, {\em breakfast}).

We then extract the posterior topic distribution for each document, along with the document term matrix, to perform SVM classification. Table~\ref{tab:topics} shows a sample of learned topics.

\begin{table*}
    \caption{All learned topics, represented by their top keywords. Those keywords occurring in negative contexts are in red.}
    \label{tab:topics}
    \centering
    \small
    \begin{tabular}{|l|l|l|}
        \hline
          order,food,\textcolor{red}{service,time,experience} & room,local,amazing,food,love & good,love,place,bake,store \\
          \hline
          food,place,just,can,\textcolor{red}{service} &   food,great,menu,drink,night & good,place,really,pretty,food\\
          \hline
          food,burrito,buffet,Chinese,place & taco,chicken,pork,try,\textcolor{red}{order} & great,food,good,place,friend\\
          \hline
          sushi,place,ayc,good,roll & coffee,place,tea,great,friend & place,poutine,food,\textcolor{red}{just},restaurant\\
          \hline
  burger,fries,good,place,like & lunch,restaurant,time,service,like & pizza,food,good,place,great\\
  \hline
  wing,good,like,night,great &  sandwich,pasta,food,fresh,place & Thai,food,curries,\textcolor{red}{order,service}\\
  \hline
  excel,food,really,great,place & meal,good,egg,breakfast,food&\\
        \hline
    \end{tabular}
    
\end{table*}

\paragraph{Combining DTM and LDA topics in SVM classification}
As discussed above, we create the DTM with TF weighting, and bind the LDA posterior topic distributions, as well as document classes (`Action' or `No Action'). As before, we optimize the number of features from the DTM within an SVM classifier. To avoid issues of imbalanced classes, we use SMOTE \citep{Chawla2002} to oversample the minority class, resulting in approximately 900 samples for each class.

The performance of different methods, in a 10-fold cross-fold validation, are shown in Table~\ref{tab:performance}.
 
\begin{table*}
    \caption{Averages (and standard deviations) of accuracy, $\kappa$, sensitivity, and specificity using 10-fold cross-validation. The $\kappa$ statistic is an unweighted measure of concordance for categorical data that measures agreement relative to what would be expected by chance \citep{kuhn2008building}. Legend: all keywords = all  6000 terms, top keywords = top 200 keywords, topics = learned 20 topics, NB = na\"{i}ve Bayes, SVM = support vector machine.}
    \label{tab:performance}
    \small
    \centering
    \begin{tabular}{|l|cccc|}
    \hline
            &accuracy& kappa & sensitivity& specificity\\
            \hline
    NB, all keywords &12.06\% (01.65\%) & 00.24\% (00.63\%) & 10.45\% (00.60\%) & 96.87\% (08.83\%)\\
    NB, top keywords  &55.81\% (04.15\%) & 08.70\%(04.66\%)  & 14.74\% (02.54\%)   & 93.76\% (02.12\%)\\
    SVM, all keywords  &  70.36\% (02.53\%) & 39.82\%(05.22\%)& 99.67\% (01.02\%) & 63.40\%(01.96\%)\\
    SVM, top keywords   &  69.18\% (02.79\%) & 41.95\%(15.29\%) & 93.36\%(19.87\%) &  66.27\%(12.05\%)\\
    SVM, all keywords + topics  & 66.98\% (02.70\%) & 32.86\%(05.57\%) & 100.00\% (0.00\%) & 60.86\% (02.04\%)\\
    SVM, topics &  78.81\%(04.33\%) & 57.75\%(08.64\%) & 74.64\% (03.91\%) & 84.27\%(05.56\%)\\
    SVM, top keywords + topics   & \textbf{90.82\% (13.46\%)} & \textbf{81.44\%(27.53\%)} & 93.46\% (07.92\%)& 90.07\%(13.46\%)\\
    \hline
    \end{tabular}
\end{table*}

\section{Conclusion} 
Textual analysis of Yelp reviews can clearly be used to predict official health inspection results, as evidenced by over 90\% accuracy in predicting official WPH actions using learned keywords and topic model features in an SVM classifier. In both the NB and SVM approaches, reducing the total number of keywords to an optimized subset provides a significant improvement in performance, most likely to possible overfitting in the full set.

Given that both sensitivity {\em and} sensitivity are very high in our optimum system, it is strongly suggested that this tool, or one very much like it, can be used by governments to optimize their resources. Specifically, inspection schedules should be automatically generated from crowd-sourced data, as in this work, and standard regulated procedures. The derivation of such schedules is the subject of future work. Other possible avenues of exploration include interactive online databases combining official health inspector reports and crowd-sourced information. Artificial intelligence is helping to shift public health informatics to the public sphere, and it will be increasingly important for that shift to incorporate new approaches to machine learning.


\bibliographystyle{apalike}

\bibliography{bib}

\end{document}